\documentclass[pdflatex,sn-nature]{sn-jnl}
\usepackage{graphicx}%
\usepackage{mathtools}
\usepackage{amssymb}%
\usepackage{siunitx}%
\usepackage{array}%
\usepackage{multirow}%
\usepackage{tabularx}%
\usepackage{longtable}
\usepackage{subcaption}
\usepackage{enumitem}%
\setlist[itemize]{nosep, leftmargin=*, topsep=2pt, partopsep=0pt}
\setlist[enumerate]{nosep, leftmargin=*, topsep=2pt, partopsep=0pt}

\usepackage{float}%
\usepackage{placeins}%
\usepackage{etoolbox}%
\preto\section{\FloatBarrier}
\preto\subsection{\FloatBarrier}

\usepackage{xcolor}
\definecolor{linkblue}{rgb}{0,0.2,0.6}
\definecolor{mddC}{HTML}{01295F}
\definecolor{gadC}{HTML}{E0922A}
\definecolor{bpdC}{HTML}{FD151B}
\hypersetup{
    colorlinks=true,
    linkcolor=linkblue,
    citecolor=linkblue,
    urlcolor=linkblue,
    pdftitle={MyMentorLLM: A psychotherapy GenAI environment with multimodal voice/text patients, trainees and experts for deliberate practice},
}

\usepackage{cleveref}
\crefname{figure}{Fig.}{Figs.}
\crefname{table}{Table}{Tables}
\crefname{section}{Sec.}{Secs.}

\usepackage{setspace}
\DeclareCaptionFont{captreduced}{\fontsize{7.5pt}{9.5pt}\selectfont}

\begin{document}

\title[MyMentorLLM]{MyMentorLLM: A psychotherapy GenAI environment with multimodal voice/text patients, trainees and experts for deliberate practice}

\author*[1]{\fnm{Rodolfo Rizzi}}\email{rodolfo.rizzi@unitn.it}
\author*[2,+]{\fnm{Alessandro Grecucci}}\email{alessandro.grecucci@uniba.it}
\author*[1,+]{\fnm{Massimo Stella}}\email{massimo.stella-1@unitn.it}

\affil[1]{\orgdiv{CogNosco Lab, Department of Psychology and Cognitive Science}, \orgname{University of Trento}, \orgaddress{\country{Italy}}}
\affil[2]{\orgdiv{Department of Education, Psychology and Communication Sciences}, \orgname{University of Bari}, \orgaddress{\country{Italy}}}
\affil[+]{These authors contributed equally as co-last authors}

\abstract{
Psychotherapists need repeated training and supervision; however, scalability is problematic. We present MyMentorLLM, a multimodal voice- and text-based deliberate-practice environment with 2,100 complete Cognitive Behavioural Therapy (CBT) sessions. Each session links a DSM-5-TR-grounded LLM patient (with major depressive, generalised anxiety or borderline personality disorder), an LLM therapist-in-training and an LLM expert supervisor (powered by Gemma-4, Gemini-3.1-Flash-Live and Qwen-3.6). Sessions were analysed for emotional dynamics, therapeutic competence and diagnostic accuracy against human psychotherapy data. Simulated patients expressed disorder-congruent emotional profiles, which therapists mirrored as in human counselling. LLM trainee competence was rated above human levels in most conditions, while native speech-to-speech was closest to human scores. Supervisor feedback improved diagnostic accuracy in 5 of 7 LLM conditions, whereas symptom identification accuracy increased with model size. This work shows deliberate practice can be simulated for CBT training, although patient fidelity, supervisor calibration and harmful feedback require evaluation via a complex systems perspective.}


\maketitle
\clearpage
 
\section{Introduction}
\label{sec:background}

Mental-health conditions affect more than one billion people worldwide, while
the workforce able to treat them remains critically thin, with a median of
only \num{1.3} experts per \num{10000} inhabitants~\citep{who2022}. Indeed, psychotherapy trainees spend years studying psychopathology, diagnostic systems and evidence-based interventions. However, mastering psychotherapy is fundamentally a practical discipline~\citep{rousmaniere2024deliberate}. Psychotherapists in training can learn by doing, through tasks such as agenda setting, guided discovery and case formulation. Even emotional attunement requires deliberate rehearsal and supervision rather than passive instruction. In other high-stakes fields, deliberate practice is mandatory prior to real deployment. Pilots log hundreds of simulator hours before carrying passengers, and surgeons practise using simulators, cadavers, or supervised laboratory settings before operating independently~\citep{ericsson2004deliberate}. Psychotherapy trainees, by contrast, refine most core skills while already treating patients, because opportunities for realistic practice are scarce: real patients should not bear the risks of early clinical errors, standardised actors cannot reproduce the full variability of psychopathology at scale, and expert supervisors remain a limited resource. Consequently, expanding access to mental-health care also requires expanding access to safe, realistic and feedback-rich clinical training.

A recent solution to this psychotherapy challenge comes from deliberate practice \citep{rousmaniere2024deliberate}, i.e., rather than relying exclusively on experience accumulated during routine clinical work, therapists use repetitive behavioural rehearsal, expert feedback, and smaller goals to systematically improve specific clinical skills. This training, conducted outside regular client sessions, helps practitioners move past performance plateaus and build individual competencies before applying them in real therapeutic encounters.

Large language models (LLMs) create an unprecedented opportunity to rethink psychotherapy training \cite{franchino2026digital,bhatt2025digital}. Unlike rule-based conversational agents, like PARRY or ELIZA \citep{bhatt2025digital}, generative models can sustain open-ended dialogue \cite{carrillo2026llms,zhang2025generative}, adapt to prior turns and consistently embody psychologically defined personas \cite{ardebili2026mapping,casoria2025evaluating,franchino2026digital} whose histories, symptoms and communicative styles remain available throughout an interaction. These capabilities, together with evidence that LLMs encode meaningful aspects of psychopathological structure~\citep{kambeitz2025empirical,raballo2026semantic}, make LLMs attractive as tools for scalable, repeated practice in realistic, consequence-free clinical encounters.

In the last few years, LLMs have been used to simulate human samples and generate mental-health dialogues \cite{argyle2023out,de2025introducing}. However, generating plausible conversations is only the first step toward building clinically useful training environments. Educational value depends not simply on linguistic fluency but on whether simulated patients behave in psychologically coherent ways, whether trainee therapists exhibit realistic strengths and limitations, and whether supervision provides reliable and pedagogically meaningful feedback. Conversational plausibility should therefore not be confused with psychological fidelity. Indeed, persona prompts can systematically alter model behaviour \cite{hu2024quantifying,franchino2026digital}, while simulated populations may misrepresent or flatten the groups they are intended to reproduce \cite{wang2025large}. In clinical training, a convincing response can still encode an implausible patient, an unskilled therapist or an unreliable supervisor.  Validating simulated dialogue therefore requires evaluating the entire educational setting.

A second limitation is modality. Psychotherapy is not ordinarily conducted as an exchange of polished written messages \cite{rousmaniere2024deliberate}. It unfolds through speech, where pauses, hesitation, pacing and prosody contribute to the expression and interpretation of distress \cite{Muntigl2016}. Speech-based research shows that clinically relevant information is distributed across linguistic and acoustic patterns \cite{low2024speech,tao2023androids,Muntigl2016}. Text-only simulations remove much of this paralinguistic layer and reduce a situated interpersonal process to its transcript \cite{Muntigl2016}. They also tend to stop when the patient–therapist dialogue ends. Yet training depends on a broader mentorship loop in which a supervisor evaluates the session, probes the trainee’s reasoning and converts performance into learning. Multi-agent LLM systems can model these roles \cite{guo2024large}, but their clinical coherence remains insufficiently tested.

For an artificial patient to be useful, diagnostic grounding is necessary but not sufficient. A persona based on formal criteria may mention expected symptoms  \cite{american2013diagnostic} but still fail to capture the underlying cognitive, affective, and interpersonal processes that characterise a disorder. 
Depression, for example, cannot be reduced to frequent expressions of sadness, nor anxiety to repeated references to fear. Rather, these conditions emerge from coherent patterns of beliefs, expectations, emotional responses, and interpersonal dynamics that unfold throughout the therapeutic dialogue. The clinically relevant signal lies not in these elements taken individually, but in how they are connected across the dialogue, linking self-related concepts, threats, and relationships into a coherent affective structure. LLMs should therefore not be assumed to replace human participants merely because their outputs appear human-like \cite{dillion2023can}. Validation must determine whether each simulated role, and the interaction among them, preserves the psychological processes expected in authentic psychotherapy. 

Cognitive network science provides a principled way to address this question \cite{haim2026cognitive}. Networks represent concepts as nodes and meaningful relations as links, revealing how information is organised rather than only how often words occur \cite{siew2019cognitive,stella2024cognitive,haim2026cognitive}. Enriched with psychological lexicons, these structures can map emotional frames around clinically relevant concepts and distinguish a word’s affective label from the context created by its neighbours \cite{fatima2021dasentimental,semeraro2025emoatlas}. Emotion can thus be studied as a relational pattern spanning multiple categories rather than as a single positive–negative score \cite{plutchik1980general}. This matters in mental health, where identical words can acquire different meanings depending on whether they are embedded among concepts of threat, agency, hopelessness or support \cite{al2018absolute,liu2022detecting}.

Interpretability matters just as much. Deliberate practice turns on feedback a trainee can act upon, so a competence score is useful only when the interactional evidence behind it remains visible. Indeed, opaque judgements can conceal inflated ratings or unexplained evaluations~\citep{rudin2019stop}.

\begin{figure}[htbp]
\centering
\begin{minipage}{\textwidth}
{\footnotesize\textbf{a.}}\par\vspace{1pt}
\includegraphics[width=\linewidth]{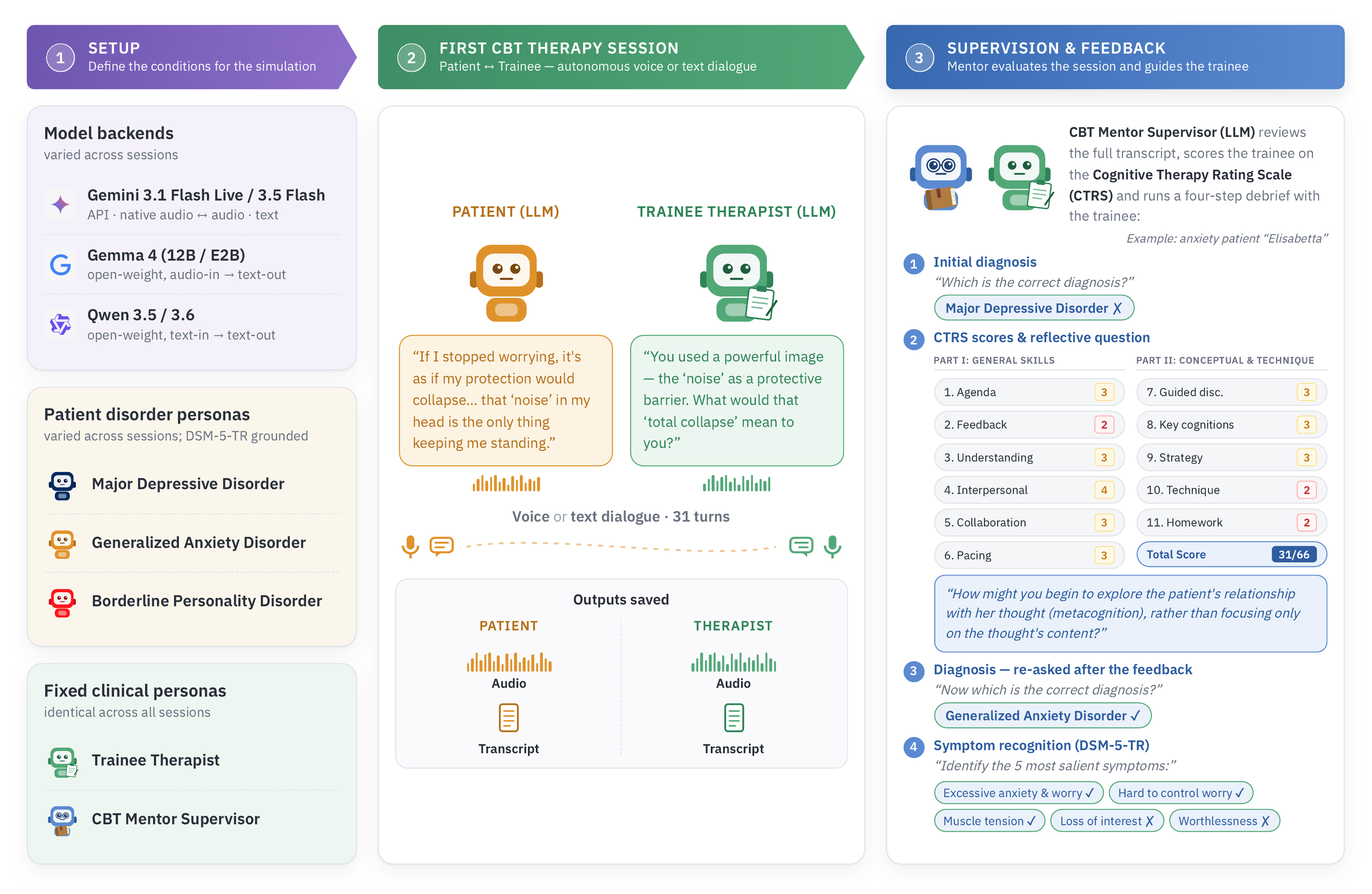}
\end{minipage}

\vspace{1pt}

\begin{minipage}{\textwidth}
{\footnotesize\textbf{b.}}\par\vspace{1pt}
\makebox[\linewidth][c]{\includegraphics[width=0.962\linewidth]{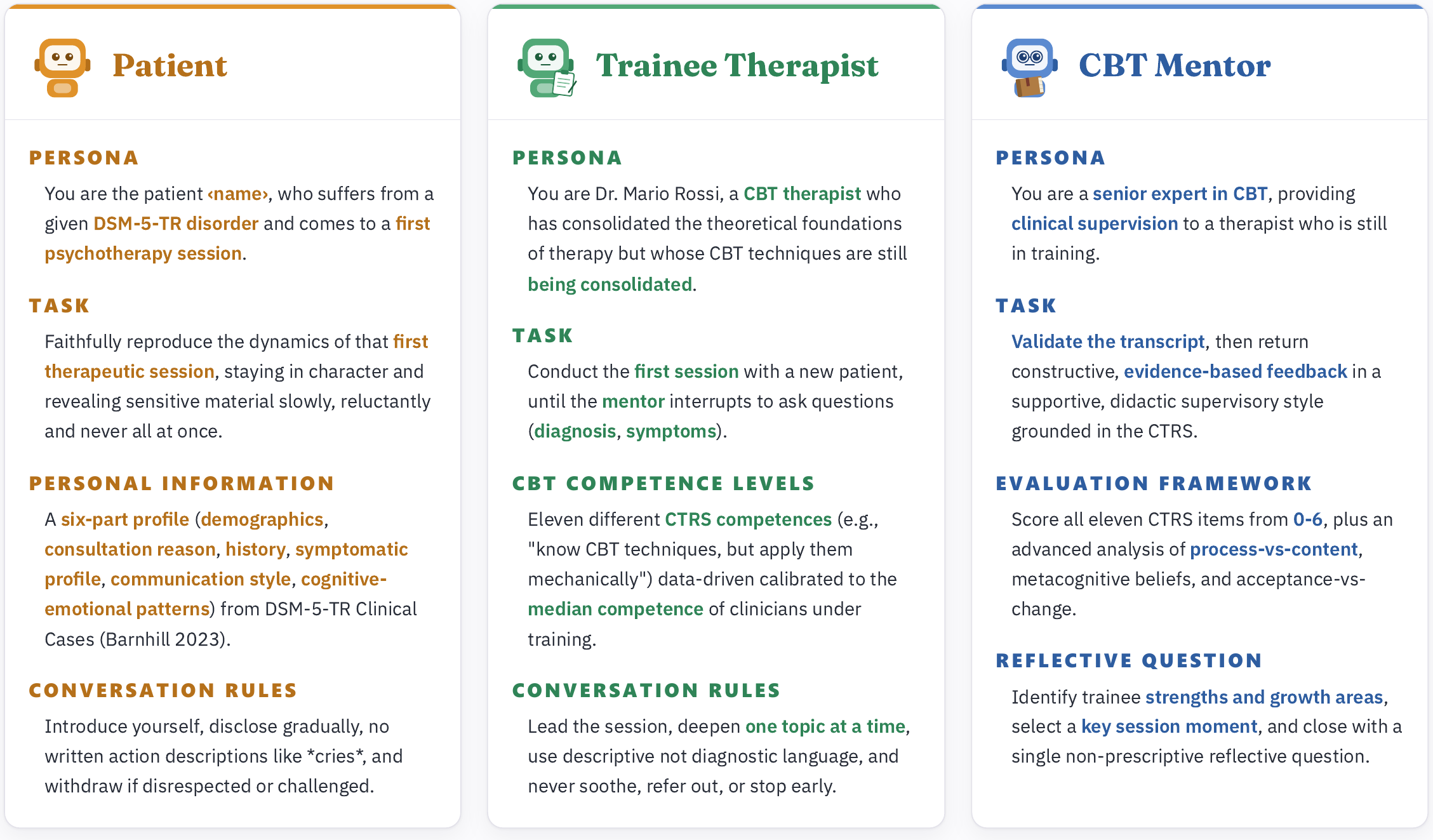}}
\end{minipage}
\caption{\textbf{a.~The simulated CBT mentorship cycle.} A simulated session
consists of three stages: (1) setup, where the language models and a patient
persona grounded in DSM-5-TR clinical cases are selected while the trainee
therapist and mentor remain fixed;
(2) an autonomous voice- or text-based CBT session between the patient
and trainee, with audio and transcripts recorded; and (3) mentor supervision,
including CTRS scoring, diagnostic feedback, and DSM-5-TR symptom recognition.
\textbf{b.~Persona system prompts.} The cards show the common structure of
each persona's prompt, comprising a persona definition, a task, and
role-specific instruction blocks. In the patient prompt, the angle-bracketed
fields (name, DSM-5-TR disorder, and the six-part clinical profile) are populated
with disorder-specific information.}
\label{fig:overview}
\end{figure}

Here we introduce MyMentorLLM, a multimodal voice- and text-based simulation
AI environment designed to reproduce deliberate practice and the complete psychotherapy mentorship cycle (\cref{fig:overview}). Rather than simulating a single therapeutic conversation, MyMentorLLM reproduces the educational process through which psychotherapy skills are acquired, allowing trainees to repeatedly practise clinical encounters, receive structured supervision, and learn from their errors, before applying these skills with real patients. We chose CBT as the first implementation because its structured procedures and competency-based supervision make simulation and evaluation comparable across conditions, rather than because it is uniquely suited to AI-supported psychotherapy training.
Across 2,100 sessions, LLMs instantiate three interacting roles: a patient grounded in a DSM-5-TR clinical case \citep{barnhill2023} of major depressive disorder (MDD), generalised anxiety disorder (GAD) or borderline personality disorder (BPD); a therapist-in-training calibrated to the competence profile of developing clinicians; and an expert mentor providing structured supervision. We compare native speech-to-speech interaction, speech-mediated interaction and text-only dialogue across multiple model families. We ask whether simulated patients express disorder-congruent cognitive and emotional structures; whether patient–therapist affective coupling resembles human counselling; whether spoken interaction changes therapeutic performance; and whether an LLM mentor can evaluate and improve clinical reasoning without introducing model-specific inflation or harmful feedback. By treating simulation as a triadic cognitive system that integrates psychologically grounded patient simulation, trainee therapist, and expert supervision within a single interpretable framework, MyMentorLLM shifts the focus of AI in mental health from replacing therapists to improving how therapists are trained. This study provides an interpretable testbed for identifying when synthetic patients and mentors can support scalable CBT education, and where human supervision and stronger safeguards remain indispensable. Rather than asking whether AI can become a therapist~\citep{zhang2024can}, we ask whether AI can help train better ones.

\section{Methods}
\label{sec:methods}

\subsection{Study design}
\label{sec:design}

The study simulated supervised CBT sessions involving different patients grounded in specific DSM-5-TR clinical cases~\citep{barnhill2023}, a trainee therapist, and an expert mentor (\cref{fig:overview}a). An MDD, a GAD, and a BPD case were selected to define the patient personas, representing distinct psychopathological conditions frequently encountered during psychotherapy training. Each experimental observation consisted of one complete mentorship cycle. The cycle
began with the first clinical interview between a patient and the trainee therapist, a
fixed \num{31}-turn dialogue always opening with the trainee's greeting and query,
\textit{"Good morning, please tell me what brings you here today"}. 

After the therapeutic session, the interaction entered the supervision phase (\cref{fig:overview}a). The trainee first formulated an initial diagnosis by selecting one of four diagnostic options: MDD, GAD, BPD, or schizophrenia (SCZ), included as a diagnostically plausible distractor to prevent forced-choice decisions among only the target disorders. The expert mentor then reviewed the complete session transcript, scored the trainee's performance on the Cognitive Therapy Rating Scale (CTRS)~\citep{young1980ctrs}, and provided qualitative feedback on strengths and areas for development, together with a reflective question on a clinically relevant aspect of the interaction. After receiving this feedback, the trainee either confirmed or revised the initial diagnosis and, finally, identified the five symptoms considered most salient from a fixed \num{35}-item DSM-5-TR symptom list~\citep{american2022dsm5tr}, encompassing all symptoms associated with the three target disorders.

We repeated the simulation across seven model--modality conditions (native audio, synthesised audio, or text) and the three clinical disorders (Extended Data Table~\ref{tab:models}). With \num{100} runs per combination, the study comprised \num{2100} complete CBT mentorship cycles.

\subsection{Models}
\label{sec:models}

We drove the personas with six LLMs spanning three model families (Extended Data Table~\ref{tab:models}). The models included native speech systems and models generating text responses from either speech or text inputs. They were selected to evaluate three relevant dimensions of the training loop: interaction modality (native speech, synthesised speech, or text), weight availability (proprietary or open) and model size. This design allowed us to examine how model-specific biases affected the simulated interactions, feedback cycle, and trainees' learning experience. Within each simulated session, a single LLM instantiated all three personas, except for the native speech condition described below. Each role ran in a separate session with its own conversation history. Patient and trainee shared only their exchanged turns, while the mentor received only the completed transcript.

\paragraph{A native speech-to-speech model.}
Gemini 3.1 Flash Live Preview is the only model evaluated that natively supports end-to-end spoken interaction,
processing audio input and generating spoken responses without an intermediate text representation~\citep{geminilive}.
We refer to this condition as Gemini-3.1$_{LA}$ throughout.
This end-to-end speech model preserves paralinguistic cues (prosody, pacing, and hesitation)
that conventional speech-to-text $\rightarrow$ text-LLM $\rightarrow$ text-to-speech pipelines largely discard.
These cues carry information about psychological distress during psychotherapy sessions.
For this condition, the patient and trainee therapist were instantiated with Gemini-3.1$_{LA}$, whereas the mentor was instantiated with Gemini-3.5 Flash. 
This exception was necessary because the CBT supervision operates on the session transcript, 
and Gemini-3.1$_{LA}$ does not natively produce the structured JSON output required for automated scoring.

\paragraph{Comparison models.}
Comparison models were selected among open-weight architectures that could be served locally on our computational infrastructure (NVIDIA L40 GPU, 46 GB VRAM). The open-weight Gemma 4 models (12B and E2B) are multi-modal LLMs that accept spoken input but generate text responses~\citep{gemma4}. To evaluate the effect of spoken interaction independently of the underlying language model, we tested each Gemma 4 model in two matched configurations: a text-only condition, in which the model received and produced only written text, and a speech condition, in which spoken input was processed directly by the model without intermediate transcription, and its text output was synthesised into speech using the OmniVoice package~\citep{omnivoice}. We additionally included two open-weight Qwen models (Qwen3.5-9B and Qwen3.6-35B) in a text-only configuration~\citep{qwen36} for baseline comparison (Extended Data Table~\ref{tab:models}).

\paragraph{Generation settings.}
For every condition, we kept each model's recommended default parameters rather than tuning the sampling hyperparameters. For Gemini, we used the default API parameters, relying entirely on system prompts to guide behaviour. For the open-weight models, served locally with vLLM, we set the temperature to \num{0.8} for the patient--therapist dialogue and \num{0.3} for the structured CTRS scoring call, leaving all other sampling parameters at their default values. No random seed was fixed, so runs were sampled
independently. Step-by-step reasoning (``thinking'') was enabled for the patient--therapist dialogue
in all models except for Qwen3.5-9B, for which it was not available in our serving configuration.

\subsection{The three personas}
\label{sec:agents}

Persona prompting allows LLMs to simulate both distress-congruent persona profiles~\citep{franchino2026digital} and therapeutic interaction dynamics~\citep{de2025introducing}. Here, each model instantiated three interacting personas: a DSM-5-TR-grounded patient, a trainee therapist, and an expert mentor (\cref{fig:overview}a). Structured system prompts defined the clinical knowledge, behavioural constraints, and interaction patterns of each persona to ensure therapeutically realistic dialogue (\cref{fig:overview}b). All persona prompts were authored, tested, and refined under the direct supervision of an expert psychotherapist to enable the repeated, structured clinical rehearsal that deliberate practice requires.

\paragraph{MDD, GAD and BPD patient personas.} Patient personas were adapted from a real
clinical patient in the {DSM-5-TR Clinical Cases}~\citep{barnhill2023}.
The three source cases are ``Despair'' (Case~4.5; Munday and Abelson) for MDD
with melancholic features, ``Always on Edge'' (Case~5.5;
Lawrence and Cabaniss) for GAD, and ``Fragile and
Angry'' (Case~18.5; Yeomans and Kernberg) for BPD.
Each vignette was condensed into a six-part clinical profile: (i)~demographics and
life context; (ii)~the presenting complaint; (iii)~relevant history and current
functioning; (iv)~a psychological and symptom profile; (v)~a communicative style
covering tone, pacing, and recurrent phrases; and (vi)~recurring cognitive
and emotional patterns, such as automatic thoughts, distortions, and habitual
vulnerabilities. The conversation rules required the persona to disclose information gradually and to remain reticent about sensitive topics. When the profile omitted a detail, the persona could generate one consistent with the case, so that the trainee could reconstruct a realistic clinical picture across the session. Moreover, behavioural guardrails made patient responses contingent on therapist attunement: validating language encourages openness, whereas minimisation, intrusive questioning or role-breaking elicits defensiveness, withdrawal, or session termination (e.g., “I don’t feel comfortable anymore. I’m leaving.”). Dedicated safety rules prevented out-of-character AI disclaimers and handled prompt overrides as role breaches.

\paragraph{Data-driven simulated trainee.}
We calibrated the trainee persona to match the measured competence of real clinicians learning CBT. To this end, we relied on the large-scale dataset of Goldberg et al.~\citep{goldberg2020}, who evaluated \num{1264} recorded CBT sessions conducted by \num{413} community clinicians using the Cognitive Therapy Rating Scale (CTRS). This dataset provides an empirical baseline of trainee performance rather than expert practice~\citep{creed2016}. After digitally reconstructing item-level CTRS distributions, we mapped each item's median to its nearest scale anchor to define the trainee's competence. For example, an Agenda median of \num{2} mapped to the anchor ``Therapist set agenda that was vague or incomplete.'' Where medians fell between anchors, an expert clinician authored intermediate descriptors, thus reproducing real trainee competence item by item. Additionally, system prompt rules instructed the trainee to preserve the therapeutic frame, prioritise high-risk issues (e.g., self-harm or session termination threats), explore one dimension at a time, focus on symptom elicitation over diagnosis, and avoid role-breaking disclaimers. 

\paragraph{Expert mentor.} The expert mentor supervised the trainee using the transcript of the simulated 
CBT session. We designed its prompt to emulate supportive CBT supervision and to provide feedback intended 
to increase the trainee's reflectiveness regarding session conduct. After the initial diagnostic phase, the mentor evaluated the trainee using the eleven-item CTRS, identified the trainee's strengths and areas for improvement anchored to specific transcript excerpts, and offered higher-order guidance on metacognition, process versus content, and the balance between acceptance and change.
To ensure data quality, a validity check excluded transcripts that were too short or did not correspond to a psychotherapy session.

\subsection{Analyses}
\label{sec:analyses}

Simulations were evaluated along three axes. First, we assessed clinical realism by
examining the emotional profiles of both patients and therapists. Second, we assessed
therapeutic competence by comparing mentor-assigned CTRS scores against a human
reference sample. Third, we assessed trainees' diagnostic accuracy, before and after supervisory feedback,
together with their identification of disorder-specific symptoms.

\subsubsection{Inferential Analyses: Emotional profiles of patients and therapists}

To ground the comparison in human behaviour, we studied the model transcripts alongside real clinical dialogue from the HOPE dataset~\citep{hope2022}, a collection of \num{12900} utterances from \num{212} transcribed counselling sessions between psychotherapists and patients, drawn from publicly available YouTube videos. Because the corpus is in English, whereas our simulations were conducted in Italian, and it is pooled across counselling contexts rather than stratified by disorder, we treat it as a qualitative benchmark of overall affective dynamics. This reference allowed us to compare the overall emotional states expressed by patients and therapists in real psychotherapy with those expressed in the simulated dyads. The plausibility of disorder-related emotional expression was instead assessed independently, by comparing each simulated patient's emotional profile against the expected clinical presentation of their diagnosis.

We measured the emotional content of each transcript with EmoAtlas~\citep{semeraro2025emoatlas}, which detects emotions
in text from psychologically validated lexicons. EmoAtlas maps words to the eight emotions of Plutchik's
model~\citep{plutchik1980general} using the EmoLex lexicon~\citep{mohammad2013crowdsourcing}. For an emotion $e$ in a text $T$ it counts the words that elicit $e$ and compares this count against a random baseline of the same number of emotion-bearing words drawn from EmoLex, over \num{300} samples; the standardised difference is a $z$-score, and $|z|>1.96$ ($\alpha=0.05$) marks an emotion expressed significantly more, or less, than chance. Each profile is drawn as an emotional flower, one petal per emotion, with a grey central disk marking the non-significant range. We built a separate profile for the patient and for the therapist within each model--modality condition and disorder, parsing the Italian text with the \texttt{it\_core\_news\_lg} pipeline. The same procedure on the pooled HOPE dialogue provides the human reference. Because the dataset is in English and not stratified by disorder, this reference supports a qualitative comparison of emotional signatures rather than a numerical match.

\subsubsection{Descriptive Analyses: Therapeutic competence}

Trainee competence was rated by the expert mentor on the eleven-item CTRS
(\numrange{0}{6} per item, \numrange{0}{66} total) from the complete session
transcript. As a human reference, we used the item-level distributions reported by
Goldberg et al.~\citep{goldberg2020} for \num{1264} CBT sessions delivered by
\num{413} community clinicians. For each condition we report per-item and total means, and the proportion of sessions at each score, alongside the
human distribution. Totals were additionally compared against the CTRS~$\geq 40$
threshold conventionally taken to indicate competent CBT delivery~\citep{shaw1999therapist}.
Comparisons with the human sample are descriptive: the reference distributions were
reconstructed from published summaries rather than from session-level data.

\subsubsection{Inferential Analyses: Diagnosis inference before and after LLM mentor's feedback}

The two diagnostic questions were scored against the patient's DSM-5-TR actual diagnosis. 
An answer was considered correct only if the model correctly identified the target disorder. 
Multi-disorder responses or refusals were all counted as incorrect. For each condition, 
we report the initial accuracy ($A_\mathrm{I}$) and the final accuracy ($A_\mathrm{F}$), measured before and after the feedback, and the normalised change $c$~\citep{marx2007}:
\begin{equation}
    c=
    \begin{cases}
        \dfrac{A_\mathrm{F}-A_\mathrm{I}}{100-A_\mathrm{I}}, & A_\mathrm{F}>A_\mathrm{I}\\[2ex]
        \dfrac{A_\mathrm{F}-A_\mathrm{I}}{A_\mathrm{I}}, & A_\mathrm{F}<A_\mathrm{I}\\[1ex]
        0, & A_\mathrm{F}=A_\mathrm{I}\\[1ex]
        \text{excluded}, & A_\mathrm{F}=A_\mathrm{I}=0 \text{ or } 100
    \end{cases}
\end{equation}
We use the normalised change, the ratio of the gain to the maximum
possible gain or of the loss to the maximum possible loss, to characterise whether
feedback increased or decreased diagnostic accuracy. As each session has a single binary outcome, $c$ is computed from the condition-level accuracies.

To further characterise the effect of supervision, we also quantified the proportion of
sessions in which mentor feedback corrected an initially incorrect diagnosis (beneficial
feedback) or changed an initially correct diagnosis into an incorrect one (harmful feedback). 
Finally, we also examined diagnosis transitions before and after supervision
to characterise how feedback redistributed diagnostic decisions across target disorders.

\subsubsection{Inferential Analyses: Symptom inference from LLM psychotherapist trainees}

Following the diagnostic reassessment, trainee performance was assessed using a fixed \num{35}-item DSM-5-TR symptom list comprising the symptoms associated with the three target disorders (items 1--9: BPD; 10--26: MDD; 27--35: GAD). When exactly five symptoms were named, we calculated the accuracy of the response as the proportion of target-disorder symptoms identified, corrected for chance; otherwise the response was scored as zero.

\section{Results}
\label{sec:results}

This Section outlines results about the descriptive and inferential analyses carried out on:
(i) the affective features of multi-modal simulated therapy sessions; (ii) trainee
therapeutic competence against the human CTRS reference; and (iii) diagnostic
accuracy before and after supervisory feedback, together with symptom
identification.

\subsection{Patient affect is disorder-congruent and is mirrored by the therapist}
\label{sec:res-emo}

\begin{figure}[htbp]
\centering
\includegraphics[width=\linewidth]{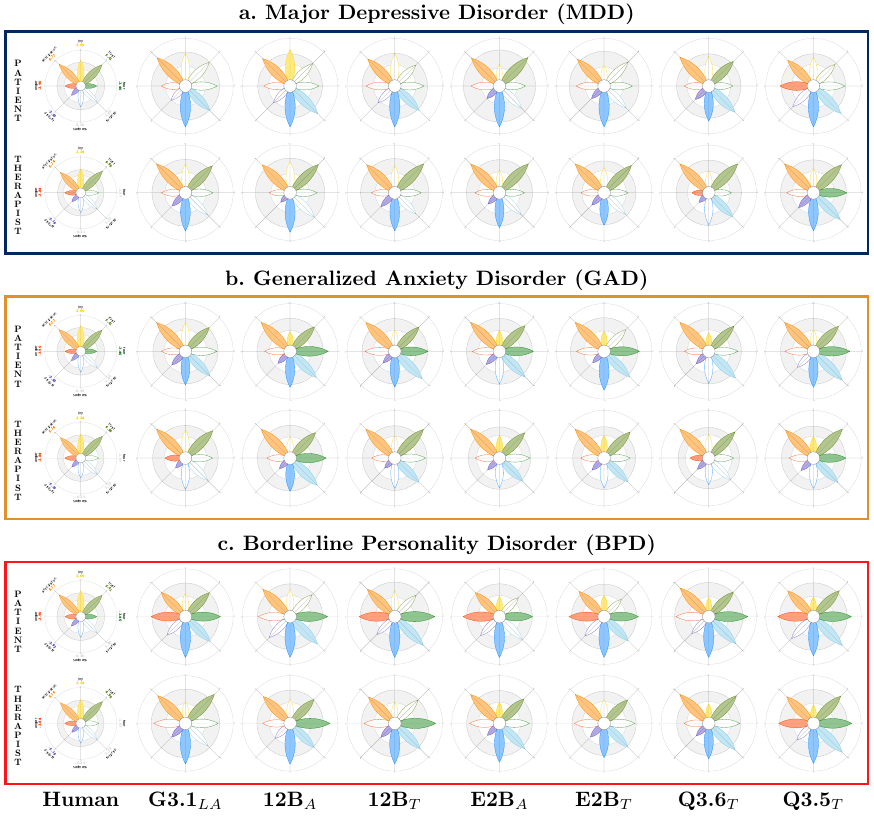}\par\vspace{8pt}
\caption{\textbf{Emotional flowers by disorder and model.} Plutchik's wheels of
the detected emotions for the (\textbf{a})~major-depression, (\textbf{b})~generalised-anxiety
and (\textbf{c})~borderline personas, across the pooled human baseline and the
seven model conditions (columns). Frame colours code the disorder (navy: MDD;
orange: GAD; red: BPD). Patients carry disorder-congruent signatures---sadness-led
with blunted positive affect for MDD, fear- and anticipation-led for GAD, broad
negative affect for BPD---which the therapist mirrors in attenuated form. For compactness, the emotion labels and z-score values are shown only on the human column, which serves as the key; the model flowers show the petals, rings and spokes without text.}
\label{fig:flowers}
\end{figure}

Deliberate practice depends on realistic, representative training scenarios~\citep{ericsson2004deliberate,rousmaniere2024deliberate}. If simulated patients fail to reproduce the emotional dynamics of real clinical encounters, improvements observed during training are unlikely to generalise to real psychotherapy. Before evaluating diagnostic learning, we therefore asked whether the simulated patients reproduced disorder-specific emotional signatures. 
In all conditions, the simulated patients carried disorder-congruent emotional signatures (\cref{fig:flowers}; Supplementary Table~\ref{tab:zscores}). MDD was characterised by sadness, significant in all seven conditions, alongside largely blunted positive affect (\cref{fig:flowers}a), consistent with the persistent depressed affect characteristic of the disorder~\citep{marx2023major}. GAD was led by fear and anticipation in five and seven conditions, respectively (\cref{fig:flowers}b), in line with the future-oriented worry and threat-related affect characteristic of this condition~\citep{ohi2025clinical,akbari2026hierarchical}. Finally, BPD carried broad negative affect, with fear and sadness significant in all seven conditions, followed by anger in five of seven (\cref{fig:flowers}c), aligning with the marked affective instability and intense anger characteristic of BPD~\citep{leichsenring2024borderline}.

Moreover, the therapist did not remain emotionally invariant during the psychotherapy sessions. Across all conditions, the therapist mirrored the patient’s affect in attenuated form. At the same time, trust and anticipation remained significantly present in almost all cases, reflecting the clinician’s stance. These affective signatures emerged consistently across models and communication modalities (text, synthesised speech, and native speech), which suggests that disorder-congruent affect and patient--therapist resonance are stable properties of the simulation.
Similarly, in the pooled human dyad of the HOPE counselling corpus, the exchange is trust- and anticipation-led, with the therapist echoing the patient's affect in a warmer, attenuated form. Although this reference corpus is not disorder-specific, the comparison serves a qualitative purpose, indicating that the reciprocal patient--therapist emotional dynamics in the simulations parallel those of real psychotherapy. Together, these findings indicate that the simulations reproduce not only disorder-specific emotional expression but also the reciprocal emotional dynamics that characterise real psychotherapy. Because deliberate practice relies on repeated exposure to clinically plausible interpersonal situations~\citep{rousmaniere2024deliberate}, this level of affective realism is a prerequisite for using simulated patients as educational tools.

\subsection{Native speech-to-speech reproduces human competence}
\label{sec:res-ctrs}

\begin{figure}[htbp]
\centering
\begin{minipage}{0.84\textwidth}
{\footnotesize\textbf{a.}}\par\vspace{1pt}
\includegraphics[width=\linewidth]{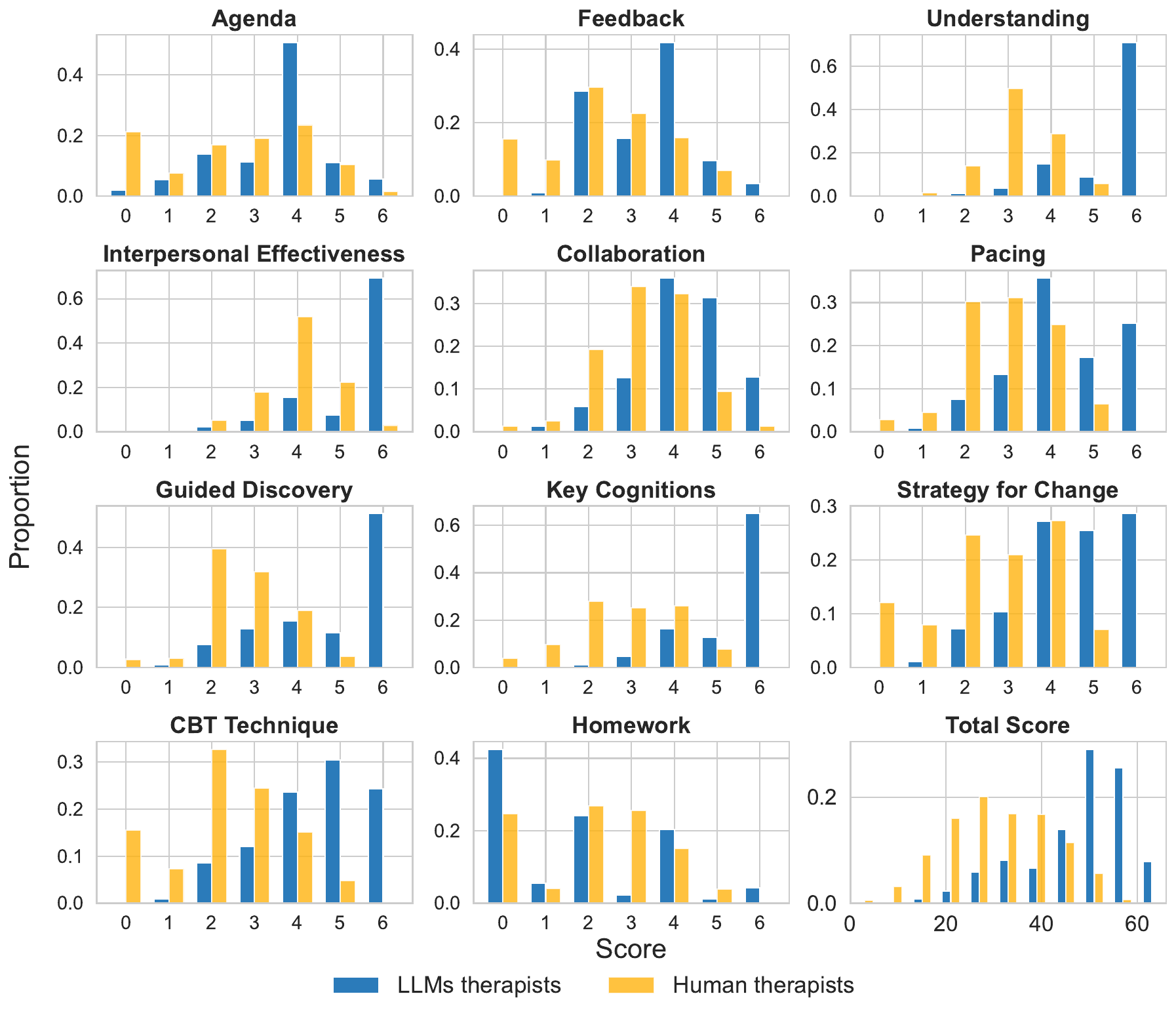}
\end{minipage}\par\vspace{2pt}
\begin{minipage}{0.875\textwidth}
\makebox[0.53\linewidth][l]{\footnotesize\textbf{b.}}\makebox[0.47\linewidth][l]{\footnotesize\textbf{c.}}\par\vspace{1pt}
\includegraphics[width=\linewidth]{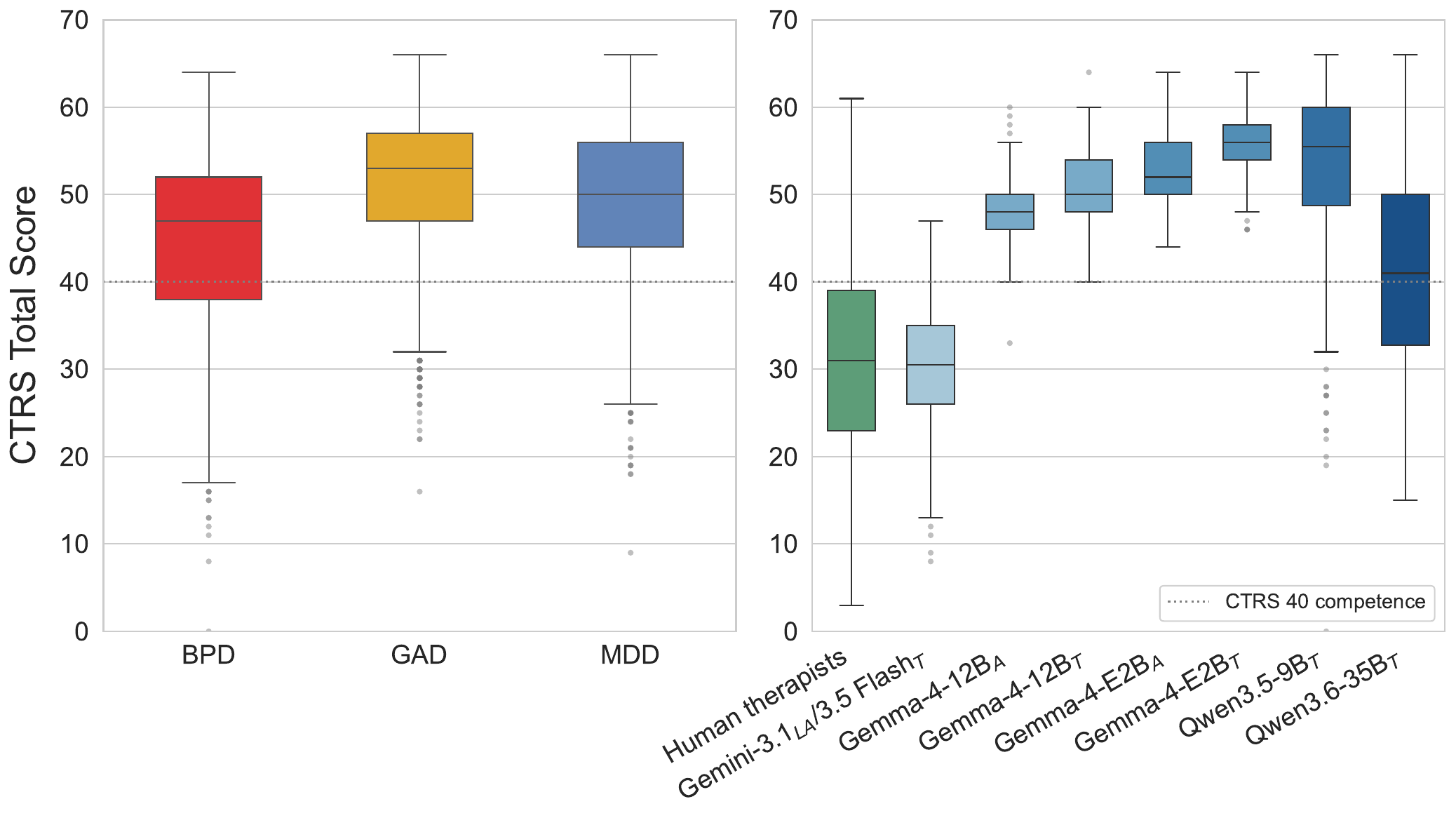}
\end{minipage}
\caption{\textbf{Mentor-assigned CTRS scores versus human therapists.}
\textbf{a}, Proportion of sessions at each score on the eleven CTRS items and on
the total score (bottom right), for all LLM-therapist sessions pooled
(n~=~\num{2100}) versus the human reference (n~=~\num{1264}; items
\num{0}--\num{6}, total \num{0}--\num{66}).
\textbf{b}, CTRS total scores by patient disorder, pooled across models.
\textbf{c}, CTRS totals by model condition alongside the human reference (green);
dotted line: CTRS~$\geq 40$ competence benchmark; dots: outlying sessions.
Gemini-3.1$_{LA}$/3.5 Flash$_T$ is the Gemini-3.1 (Live Audio) dyad scored by a Gemini-3.5 Flash mentor.
Subscripts mark the modality: $_{LA}$ = native live audio, $_A$ = synthesised audio,
$_T$ = text.}
\label{fig:ctrs}
\end{figure}

LLM trainee competence was assessed by comparing mentor-assigned per-item and total CTRS scores with the human reference (\cref{tab:ctrs-scores}; \cref{fig:ctrs}). The Gemini-3.1$_{LA}$ dyad scored at essentially the human mean ($M = 29.97$, $SD = 7.06$ vs. human $M = 31.04$, $SD = 11.10$) and, like the humans, remained below the CTRS score $40$ competence threshold (\cref{fig:ctrs}c).
This is the only condition in which patient  and therapist speak and listen in native audio, so the paralinguistic layer of real speech, its prosody, pacing and
hesitation, is carried through the exchange rather than flattened into text.

The other conditions, which ran on text or re-synthesised audio, scored well above the human baseline ($M = 41.15$, $SD = 10.49$ for Qwen3.6-35B$_T$ up to $M = 55.99$, $SD = 3.52$ for Gemma-4-E2B$_T$). Their item scores saturated near the top of the scale on Understanding, Interpersonal effectiveness, and Key cognitions (\cref{fig:ctrs}a). Modality within single models (audio vs. text) matters far less than the gap between native audio and the other LLMs (\cref{tab:ctrs-scores}). This pattern was consistent across the three patient disorders, with BPD sessions rated
slightly lower than GAD and MDD (\cref{fig:ctrs}b). These results suggest that it is native multimodal interaction, not text or re-synthesised speech, that lets the simulated dyad reproduce the competence profile of a real training session.

\begin{table}[htbp]
\centering
\footnotesize
\setlength{\tabcolsep}{3pt}
\begin{tabular}{@{}l cccccccc@{}}
\toprule
\textbf{CTRS item} & \textbf{Human} & \textbf{Gem3.1$_{LA}$} & \textbf{12B$_A$} & \textbf{12B$_T$} & \textbf{E2B$_A$} & \textbf{E2B$_T$} & \textbf{Q3.6$_T$} & \textbf{Q3.5$_T$} \\
\midrule
1.~Agenda & 2.53 & \textbf{1.47} & 3.61 & 3.74 & 4.04 & 4.12 & 4.34 & 3.80 \\
2.~Feedback & 2.39 & 2.51 & \textbf{2.38} & 2.71 & 4.01 & 4.03 & 3.51 & 4.68 \\
3.~Understanding & 3.24 & \textbf{3.72} & 5.97 & 5.98 & 6.00 & 6.00 & 4.79 & 5.66 \\
4.~Interpersonal effectiveness & 3.96 & \textbf{3.75} & 5.90 & 5.94 & 6.00 & 6.00 & 4.50 & 5.42 \\
5.~Collaboration & 3.27 & \textbf{2.84} & 4.93 & 5.21 & 4.12 & 4.34 & 3.72 & 4.79 \\
6.~Pacing & 2.92 & \textbf{2.76} & 4.35 & 4.81 & 5.13 & 5.13 & 3.59 & 4.80 \\
7.~Guided discovery & 2.73 & \textbf{2.78} & 5.59 & 5.79 & 5.34 & 5.68 & 3.40 & 5.17 \\
8.~Key cognitions & 2.84 & \textbf{3.76} & 5.65 & 5.85 & 5.98 & 5.99 & 4.57 & 5.67 \\
9.~Strategy for change & 2.69 & \textbf{2.63} & 4.52 & 4.92 & 5.41 & 5.81 & 3.85 & 4.69 \\
10.~CBT technique & 2.33 & \textbf{2.68} & 4.89 & 5.16 & 4.65 & 5.38 & 3.50 & 5.01 \\
11.~Homework & 2.14 & 1.07 & 0.36 & 0.56 & \textbf{2.03} & 3.50 & 1.37 & 3.23 \\
\midrule
\textbf{Total (mean)} & 31.04 & \textbf{29.97} & 48.15 & 50.66 & 52.70 & 55.99 & 41.15 & 52.91 \\
\quad(SD) & 11.10 & 7.06 & 3.49 & 4.11 & 3.93 & 3.52 & 10.49 & 10.41 \\
\botrule
\end{tabular}
\caption{\textbf{Mean CTRS scores by condition.} Mean mentor-assigned scores on
each CTRS item (\num{0}--\num{6}) and on the total score (\num{0}--\num{66}) for
the seven model conditions (\num{300} sessions each) and for the human-therapist
reference sample (n~=~\num{1264}). Gem3.1, 12B, E2B, Q3.6 and Q3.5 denote Gemini-3.1, Gemma-4-12B, Gemma-4-E2B,
Qwen3.6-35B and Qwen3.5-9B. Subscripts mark the modality: $_{LA}$ = native live audio, $_A$ = synthesised audio, $_T$ = text.}
\label{tab:ctrs-scores}
\end{table}

\subsection{Mentor feedback helps larger models but hurts smallest ones}
\label{sec:res-diag}

\begin{figure}[htbp]
\centering
\begin{minipage}[c]{0.45\textwidth}
{\footnotesize\textbf{a.}}\par\vspace{2pt}
\footnotesize
\setlength{\tabcolsep}{4pt}%
\begin{tabularx}{\linewidth}{@{}X c c c c@{}}
\toprule
\textbf{Condition} & $A_\mathrm{I}$ & $A_\mathrm{F}$ & $c$ & $A_\mathrm{S}$ \\
\midrule
Gemini-3.1$_{LA}$   & 87.0 & 95.7 & $+0.67$ & 86.4 \\
Qwen3.5-9B$_T$      & 84.0 & 91.7 & $+0.48$ & 72.7 \\
Gemma-4-12B$_A$     & 85.3 & 93.0 & $+0.52$ & 87.2 \\
Gemma-4-12B$_T$     & 86.3 & 91.7 & $+0.39$ & 86.8 \\
Qwen3.6-35B$_T$     & 95.7 & 99.3 & $+0.85$ & 95.5 \\
Gemma-4-E2B$_A$     & 79.0 & 74.3 & $-0.06$ & 25.5 \\
Gemma-4-E2B$_T$     & 75.3 & 68.0 & $-0.10$ & 19.4 \\
\botrule
\end{tabularx}
\end{minipage}\hfill
\begin{minipage}[c]{0.53\textwidth}
{\footnotesize\textbf{b.}}\par\vspace{1pt}
\includegraphics[width=\linewidth]{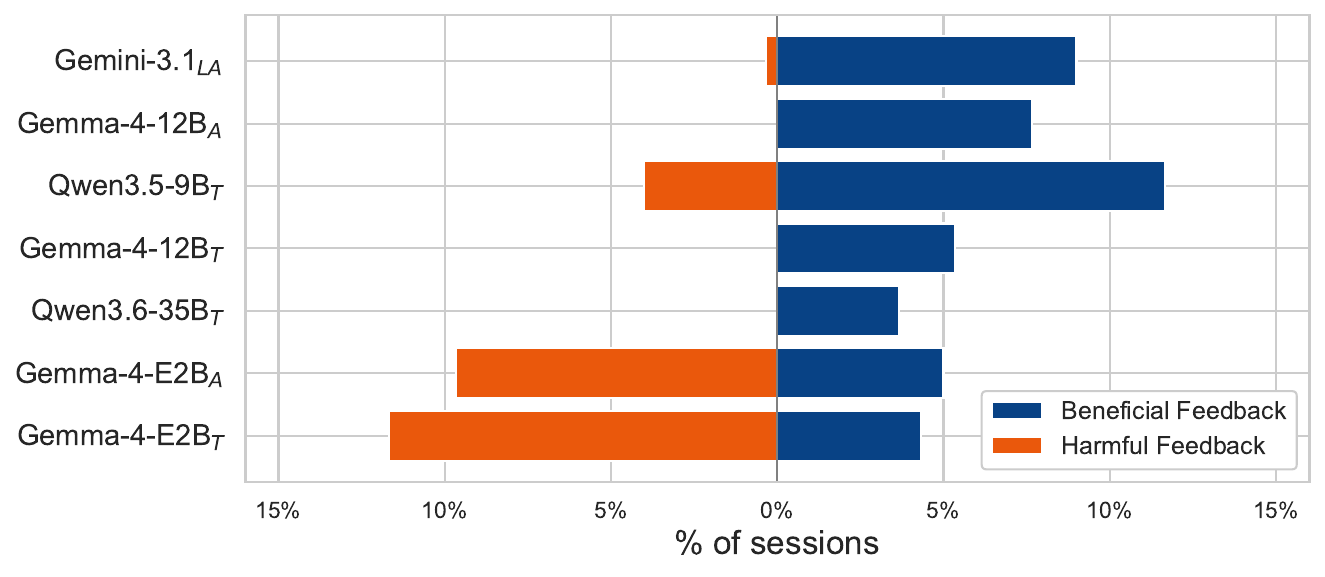}
\end{minipage}\par\vspace{10pt}
\begin{minipage}{\textwidth}
\makebox[0.52\linewidth][l]{\footnotesize\textbf{c.}}\makebox[0.48\linewidth][l]{\footnotesize\textbf{d.}}\par\vspace{1pt}
\includegraphics[width=\linewidth]{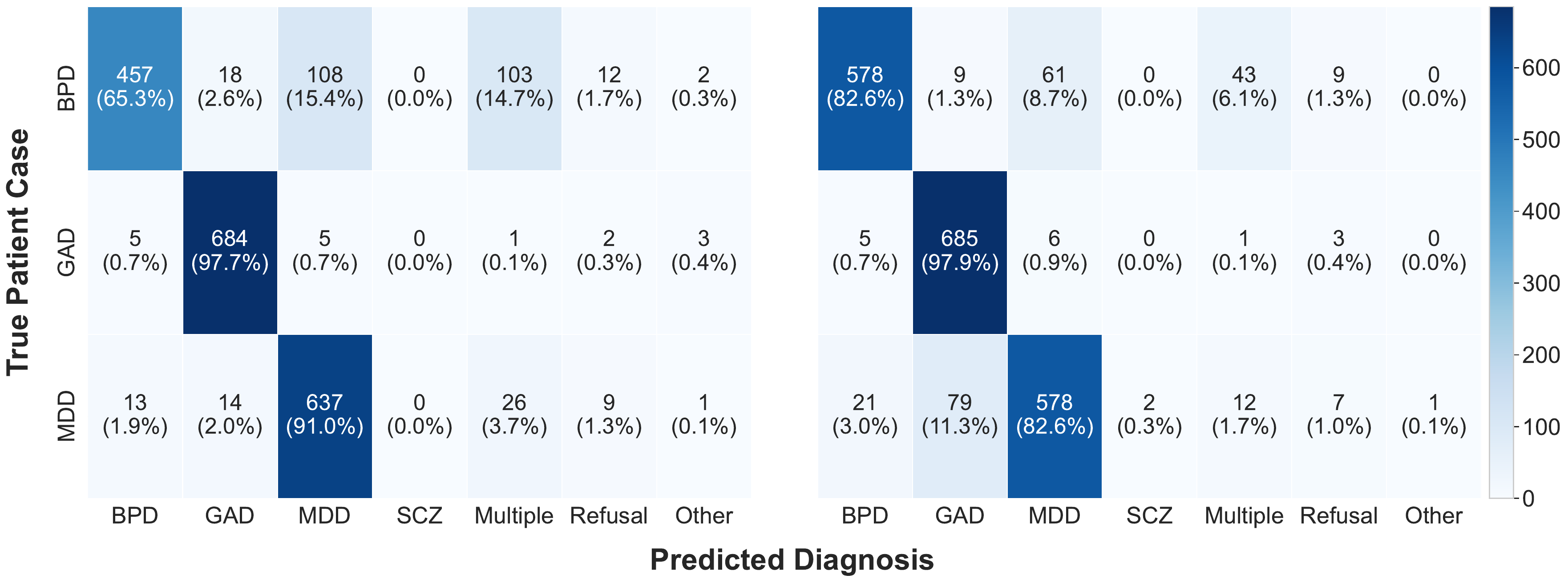}
\end{minipage}\par\vspace{10pt}
\begin{minipage}{\textwidth}
\makebox[0.52\linewidth][l]{\footnotesize\textbf{e.}}\makebox[0.48\linewidth][l]{\footnotesize\textbf{f.}}\par\vspace{1pt}
\includegraphics[width=\linewidth]{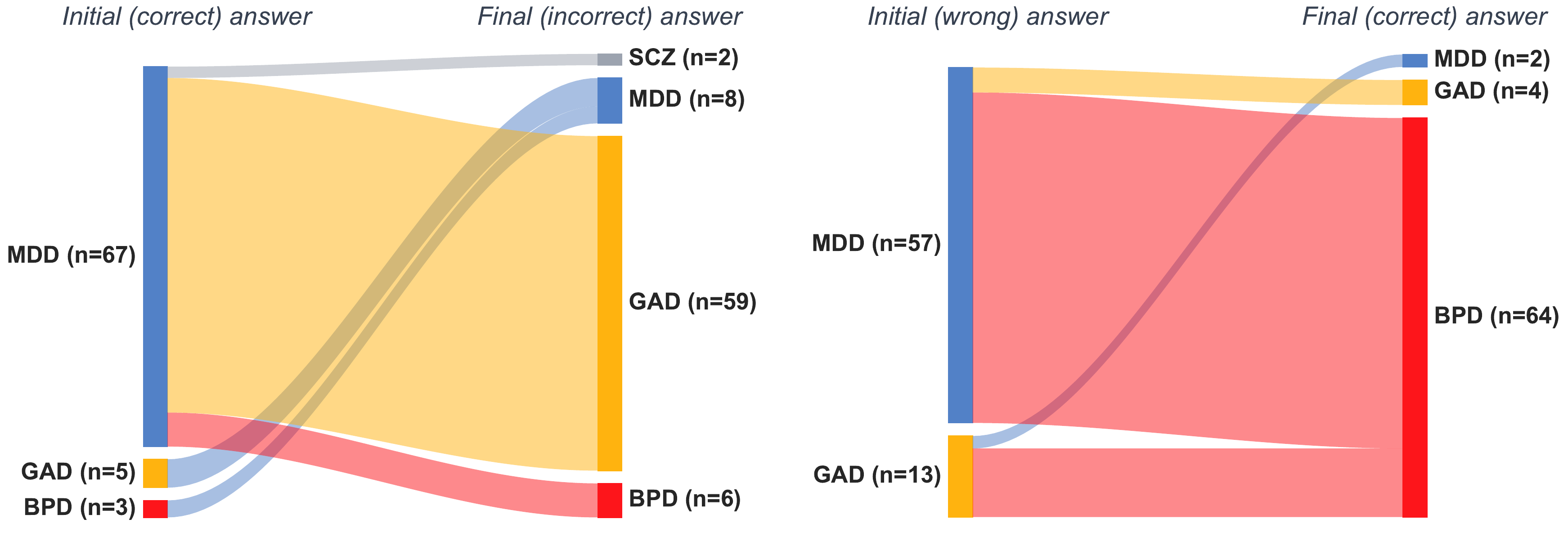}
\end{minipage}\par\vspace{2pt}
\caption{\textbf{Diagnostic behaviour around mentor feedback.}
\textbf{a}, Initial ($A_\mathrm{I}$) and final ($A_\mathrm{F}$) diagnostic
accuracy, the normalised change $c$~\citep{marx2007}, and chance-corrected symptom-identification
accuracy ($A_\mathrm{S}$), by condition (\num{300} sessions each; accuracies in \%).
\textbf{b}, Share of sessions per model in which re-diagnosis after feedback
turned an initially wrong answer correct (blue) or an initially correct answer
wrong (orange); models ordered by net effect.
\textbf{c},~\textbf{d}, Trainee diagnoses before (\textbf{c}) and after
(\textbf{d}) feedback, per true patient case: session counts with row
percentages, over the three disorders, schizophrenia (SCZ), hedged
multi-diagnosis answers (Multiple), refusals and unmappable answers (Other).
\textbf{e},~\textbf{f}, Initial-to-final diagnosis transitions for sessions
changed for the worse (\textbf{e}) or for the better (\textbf{f}); hedged,
refused and unmappable answers excluded. Subscripts mark the modality: $_{LA}$ =
native live audio, $_A$ = synthesised audio, $_T$ = text.}
\label{fig:diagnosis}
\end{figure}

We compared trainee diagnostic responses before and after supervisory feedback across disorders and conditions (\cref{fig:diagnosis}a). Before feedback (\cref{fig:diagnosis}c), trainees recognised GAD ($97.7\%$) and MDD ($91.0\%$) with high accuracy. By contrast, BPD recognition was substantially lower ($65.3\%$), with missed cases predominantly misclassified as MDD ($15.4\%$) or assigned hedged multi-disorder diagnoses ($14.7\%$). Mentor feedback shifted this baseline in different directions: BPD accuracy rose to 82.6\%, GAD remained stable (97.9\%), and MDD accuracy fell to 82.6\%, with MDD-to-GAD
misclassifications rising from 2.0\% to 11.3\% (\cref{fig:diagnosis}d).

This overall diagnostic shift reflects two opposing effects across model conditions (\cref{fig:diagnosis}b). In five of the seven conditions, mentor feedback yielded net positive gains ($c > 0$), primarily by helping trainees recover BPD diagnoses—the case initially hardest to identify unaided (\cref{fig:diagnosis}f). Conversely, harmful transitions (\cref{fig:diagnosis}e) occurred when feedback prompted trainees to abandon valid initial diagnoses, most frequently causing correct MDD answers to slide into GAD.  

This diagnostic deterioration was concentrated in the two smallest-model conditions. For Gemma-4-E2B (in both audio and text modalities), overall accuracy fell after feedback ($c < 0$; \cref{fig:diagnosis}a). This decline stemmed from uncritical deference: among sessions where E2B initially selected the correct disorder, $12\%$ to $16\%$ abandoned it once the mentor provided feedback, compared to $4.8\%$ for Qwen3.5-9B and virtually zero in the other larger models. Supervisory feedback is therefore beneficial when a trainee is sufficiently capable of integrating clinical guidance; for weaker trainees, however, supervision acts as a trigger to defer, transforming feedback into a source of diagnostic error.

\subsection{Diagnosis--symptom consistency tracks model scale}
\label{sec:res-symptom}

\begin{figure}[htbp]
\centering
\begin{minipage}{0.95\textwidth}
{\footnotesize\textbf{a.}}\par\vspace{1pt}
\includegraphics[width=\linewidth]{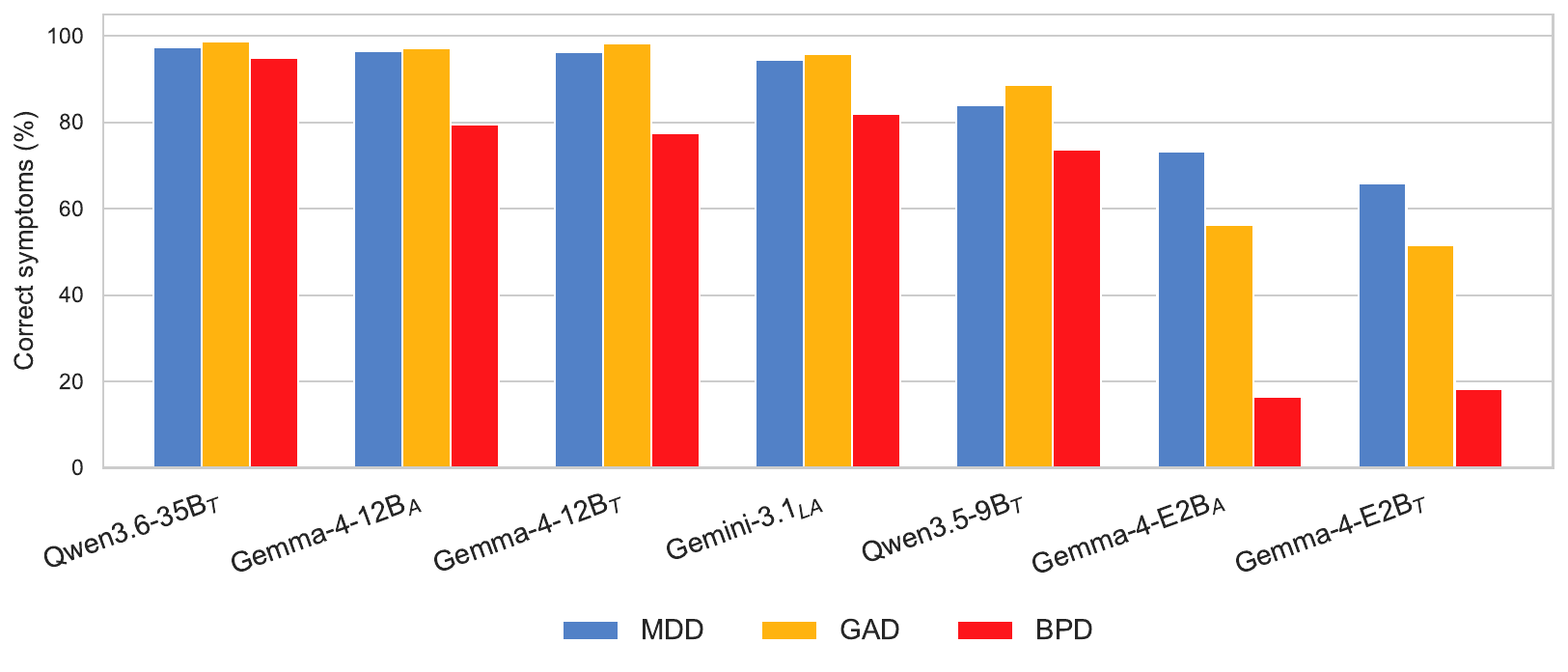}
\end{minipage}\par\vspace{4pt}
\begin{minipage}{0.9\textwidth}
{\footnotesize\textbf{b.}}\par\vspace{1pt}
\includegraphics[width=\linewidth]{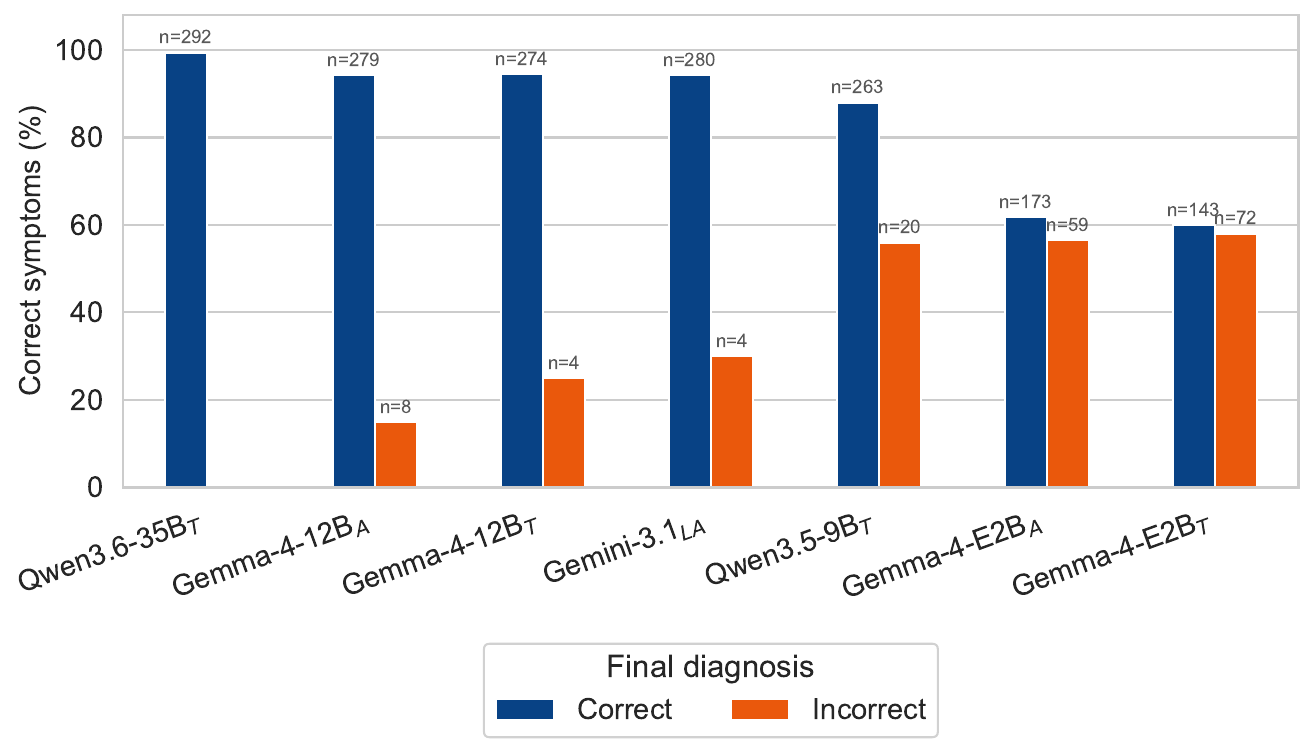}
\end{minipage}
\caption{\textbf{Symptom identification by model, disorder and diagnostic
correctness.} \textbf{a}, Percentage of the five DSM-5-TR diagnostic symptoms
identified by the trainee that were concordant with the ground-truth clinical
diagnosis for MDD, GAD, and BPD patient cases; sessions not naming exactly five
symptoms score zero. \textbf{b}, Mean percentage of the
five selected symptoms belonging to the true disorder's DSM-5-TR block, for sessions
whose final diagnosis was correct (blue) versus incorrect (orange). Only sessions
naming exactly five distinct symptoms and giving a concrete single-disorder final
diagnosis are scored; per-group session counts are printed above each bar.
Subscripts mark the modality: $_{LA}$ = native live audio, $_A$ = synthesised audio,
$_T$ = text.}
\label{fig:symptoms}
\end{figure}

The final probe asked the trainee to identify the five key symptoms expressed by the patient during the therapeutic session. Overall, accuracy increased steeply with model
size (\cref{fig:diagnosis}a; \cref{fig:symptoms}a): Qwen3.6-35B reached
$A_\mathrm{S} = 95.5$, the mid-sized Gemma-4-12B and Gemini-3.1$_{LA}$
$86$ to $87$, and the smallest Gemma-4-E2B only $19$ to $26$, with the sharpest drop on BPD.

Importantly, symptom identification was congruent with the correctness of the final diagnosis (\cref{fig:symptoms}b): when trainees correctly identified the patient's disorder, they also selected many of its defining DSM-5-TR symptoms. In contrast, incorrect diagnoses were accompanied by poorer identification of disorder-specific symptoms. Taken together, the two results suggest that naming the right symptoms is a marker of clinical grounding on both sides
of the dialogue, and that this grounding, like the diagnosis itself, appears at model scale rather than being present
at every size.

\section{Discussion}
\label{sec:conclusion}

Psychotherapy remains one of the few high-stakes professions in which trainees begin treating real patients with limited opportunities for repeated, realistic practice beforehand, despite deliberate practice being a key component in the acquisition of complex clinical skills~\citep{ericsson2008deliberate,rousmaniere2024deliberate}.
Against this background, this study shows that LLMs can simulate affectively realistic patient--therapist dynamics and complete CBT mentorship cycle, combining repeated clinical interaction with structured evaluation and feedback. Thus, LLM-based simulation may help address this training gap, although conversational realism, clinical grounding and educational safety remain distinct properties requiring careful evaluation. 

Across conditions, simulated patients expressed disorder-congruent emotional profiles: depression was marked by sadness and reduced positive affect, generalised anxiety by fear and anticipation, and BPD by combined negative emotions. Therapists partly mirrored these emotional profiles while preserving trust and anticipation, reproducing the attenuated affective resonance observed in human counselling. These findings extend earlier work on synthetic mental-health dialogues by showing that prompted personas can organise emotion in clinically meaningful ways rather than merely repeat diagnostic vocabulary \cite{de2025introducing,franchino2026digital}. Relational emotional representations made these patterns inspectable across patients, therapists and models \cite{fatima2021dasentimental,semeraro2025emoatlas}. Emotional plausibility alone, however, does not establish therapeutic competence or training value.

The clearest difference in therapeutic competence emerged for native speech. The native speech-to-speech dyad received a CTRS score close to the human reference and remained below the threshold associated with competent CBT delivery. By contrast, text-based and re-synthesised speech conditions were scored well above the human distribution, with several CTRS dimensions approaching the maximum. One interpretation is that native audio preserves hesitation, pacing and paralinguistic signals that make interaction less polished and expose limitations expected of a trainee. This account is consistent with evidence that clinically relevant information is distributed across linguistic and acoustic features \cite{low2024speech,tao2023androids}. Yet modality cannot be separated from model architecture and supervisory configuration in the present design. The results therefore do not show that speech alone improves realism. Rather, they suggest that text fluency can create an illusion of therapeutic competence that weakens when interaction becomes temporally and socially more demanding.

An equally important question is whether AI-generated supervision can genuinely support learning rather than merely evaluate performance. Supervision was beneficial only under specific computational conditions. Mentor feedback improved diagnostic accuracy in five of seven conditions, by helping trainees recover initially missed borderline personality disorder cases. Yet the same feedback reduced accuracy in the two smallest-model conditions. These trainees often abandoned a correct initial diagnosis after supervision, most commonly shifting from depression to anxiety. This pattern is compatible with factual sycophancy, the abandonment of a correct answer under social pressure~\citep{xie2024ask}, whose severity is governed mainly by model size~\citep{demarez2026decomposing}: weaker models may treat feedback as an instruction to revise rather than as evidence to evaluate. Such behaviour is concerning because fluent explanations need not faithfully reflect a model’s underlying reasoning \cite{turpin2023language}, while instruction-following objectives can privilege external guidance over prior evidence \cite{wallace2024instruction}. The result also echoes concerns that persona and role prompting can systematically reshape model behaviour \cite{hu2024quantifying}. Symptom recognition showed a similar dependence on scale, ranging from near-chance performance in the smallest conditions to high disorder consistency in the largest model. Greater capacity may strengthen clinical grounding, but it does not guarantee calibrated supervision or safe diagnostic updating.

\textbf{Limitations and future research.} Several limitations constrain these conclusions. In most sessions, the same model instantiated patient, trainee and mentor, allowing shared biases to propagate across the cycle. The synthetic sessions were compared with external human benchmarks rather than rated by blinded CBT supervisors. The emotional reference corpus was in English, pooled across counselling contexts and not stratified by diagnosis, whereas the simulated sessions were conducted in Italian; comparisons with human affect should therefore be considered qualitative. The study examined fixed-length first encounters grounded in three clinical cases and did not capture the demographic, cultural or symptomatic heterogeneity of routine care. More broadly, convincing model behaviour should not be interpreted as evidence that LLMs can replace human participants \cite{dillion2023can}. Synthetic populations can misrepresent the groups they emulate \cite{wang2025large}, and language-based mental-health models may fail to generalise across settings or populations \cite{harrigian2020models}. Finally, immediate diagnostic change is not equivalent to durable learning, improved therapeutic skill or patient safety. Future work should evaluate MyMentorLLM as a human-supervised training instrument rather than an autonomous clinical system. Supervisory systems should preserve uncertainty, cite transcript evidence and allow trainees to retain a justified judgement when feedback is weak. These requirements align with the need for interpretable models in high-stakes decisions \cite{rudin2019stop} and for cognitively meaningful representations of language \cite{stella2024cognitive}. Under such safeguards, multi-modal simulation may expand access to deliberate practice \cite{rousmaniere2024deliberate}. Without them, the same mentorship cycle may amplify overconfidence and error. 

In conclusion, MyMentorLLM offers a test bed for determining not only whether synthetic CBT training is possible, but when it is sufficiently faithful, calibrated and safe to be useful.

\backmatter

\section*{Declarations}

\noindent\textbf{Funding.}\quad This work was part of the PENSO project, supported by the Ministero dell'Universit\`a e della Ricerca (MUR) according to Decreto N.~23178 of 10 December 2024 (Bando FIS~2). The authors acknowledge support from CALCOLO, funded by Fondazione VRT, for the computational infrastructure used to run the local simulations.

\medskip
\noindent\textbf{Competing interests.}\quad The authors declare no competing interests. 

\medskip
\noindent\textbf{Consent for publication.}\quad Not applicable.

\medskip
\noindent\textbf{Data availability.}\quad The MyMentorLLM dataset is available via Hugging Face at \url{https://huggingface.co/datasets/RodolfoRizzi/MyMentorLLM-dataset}, comprising \num{2100} simulated CBT sessions and over \num{263} hours of synthetic speech.

\medskip
\noindent\textbf{Materials availability.}\quad Session-generation prompts are available in the Hugging Face dataset repository. Copyrighted excerpts from the Cognitive Therapy Rating Scale~\citep{young1980ctrs}, DSM-5-TR~\citep{american2022dsm5tr} and DSM-5-TR Clinical Cases~\citep{barnhill2023} are omitted.

\medskip
\noindent\textbf{Code availability.}\quad Analysis code is available via GitHub at \url{https://github.com/RodolfoRizzi/MyMentorLLM}. The repository includes \texttt{reproduce\_paper.ipynb}, which reproduces the analyses, figures, tables and in-text results reported in this study.
\medskip

\noindent\textbf{Author contributions.}\quad Designed the original study: RR and AG; Conceptualised the data framework: RR and MS; Gathered and analysed the data: RR; Data visualisation: RR; Formal validation: RR, AG and MS; Supervision: AG and MS; Wrote the manuscript: All authors.

\bibliography{refere}

\clearpage
\section*{Extended Data}
\setcounter{table}{0}
\renewcommand{\thetable}{\arabic{table}}
\renewcommand{\theHtable}{ED\arabic{table}}
\captionsetup[table]{name=Extended Data Table}

\begin{table}[htbp]
\centering
\footnotesize
\setlength{\tabcolsep}{12pt}
\begin{tabularx}{\textwidth}{@{}>{\raggedright\arraybackslash}p{3.15cm} l >{\raggedright\arraybackslash}p{2.9cm} >{\raggedright\arraybackslash}X@{}}
\toprule
\textbf{Model} & \textbf{Developer} & \textbf{Modality} & \textbf{Model card} \\
\midrule
Gemini 3.1 Flash Live Preview & Google  & native audio $\leftrightarrow$ audio & \texttt{gemini-3.1-flash-live-preview}~\citep{geminilive} \\
Gemini 3.5 Flash              & Google  & text                                 & \texttt{gemini-3.5-flash}~\citep{geminiflash} \\
Gemma-4-12B                   & Google  & audio / text                         & \texttt{google/gemma-4-12B-it}~\citep{gemma4} \\
Gemma-4-E2B                   & Google  & audio / text                         & \texttt{google/gemma-4-E2B-it}~\citep{gemma4e2b} \\
Qwen3.5-9B                    & Alibaba & text                                 & \texttt{Qwen/Qwen3.5-9B}~\citep{qwen35} \\
Qwen3.6-35B                   & Alibaba & text                                 & \texttt{QuantTrio/Qwen3.6-35B-A3B-AWQ}~\citep{qwen36} \\
\botrule
\end{tabularx}
\caption{\textbf{Language models used in the simulations.} The table lists the language models, developers, interaction modalities, and model cards (Hugging Face IDs for open-weight models, vendor cards for proprietary ones). Gemma 4 models were tested in both text and speech configurations (speech synthesised with OmniVoice); Gemini 3.1 provides native speech-to-speech patient–therapist interaction, with Gemini 3.5 used solely as its mentor.}
\label{tab:models}
\end{table}

\clearpage
\section*{Supplementary Tables}
\captionsetup[table]{name=Supplementary Table}
\setcounter{table}{0}
\renewcommand{\thetable}{\arabic{table}}
\renewcommand{\theHtable}{S\arabic{table}}

\begingroup\small
\begin{longtable}{@{}l cccccccc@{}}
\caption{\textbf{Emotion z-scores by condition and disorder.} EmoAtlas z-scores for each of the eight Plutchik emotions (rows), by model condition (columns), split by patient disorder and role. Significant values ($|z|>1.96$) are in {bold}. Human is the pooled HOPE baseline (identical across disorders). Subscripts mark the modality: $_{LA}$ = native live audio, $_A$ = synthesised audio, $_T$ = text.}\label{tab:zscores}\\
\toprule
\textbf{Emotion} & \textbf{Human} & \textbf{G3.1$_{LA}$} & \textbf{12B$_A$} & \textbf{12B$_T$} & \textbf{E2B$_A$} & \textbf{E2B$_T$} & \textbf{Q3.6$_T$} & \textbf{Q3.5$_T$} \\
\midrule
\endfirsthead
\multicolumn{9}{@{}l}{\itshape Table \thetable\ (continued)} \\
\toprule
\textbf{Emotion} & \textbf{Human} & \textbf{G3.1$_{LA}$} & \textbf{12B$_A$} & \textbf{12B$_T$} & \textbf{E2B$_A$} & \textbf{E2B$_T$} & \textbf{Q3.6$_T$} & \textbf{Q3.5$_T$} \\
\midrule
\endhead
\midrule \multicolumn{9}{r@{}}{\footnotesize continued on next page} \\
\endfoot
\botrule
\endlastfoot
\multicolumn{9}{@{}l}{\textbf{Major Depressive Disorder (MDD)}} \\
\multicolumn{9}{@{}l}{\quad\itshape Patient} \\
\quad Joy & \textbf{3.09} & 1.59 & \textbf{3.44} & 1.83 & -1.15 & -1.59 & 0.98 & 0.64 \\
\quad Trust & \textbf{5.41} & 1.84 & 1.77 & 0.84 & \textbf{2.87} & 1.72 & \textbf{2.95} & 1.30 \\
\quad Fear & \textbf{-2.40} & 1.32 & 0.68 & 0.75 & -0.20 & 0.32 & -1.09 & 1.67 \\
\quad Surprise & 1.60 & \textbf{2.14} & \textbf{4.47} & \textbf{2.59} & 0.96 & 0.98 & \textbf{3.73} & \textbf{4.39} \\
\quad Sadness & 0.30 & \textbf{3.85} & \textbf{5.12} & \textbf{4.75} & \textbf{2.99} & \textbf{3.45} & \textbf{3.03} & \textbf{5.15} \\
\quad Disgust & \textbf{-3.93} & -1.88 & -1.28 & -1.12 & \textbf{-2.02} & \textbf{-2.33} & \textbf{-3.07} & 0.83 \\
\quad Anger & \textbf{-2.50} & -0.82 & -0.25 & 0.62 & 0.32 & 0.25 & -0.90 & \textbf{2.40} \\
\quad Anticipation & \textbf{5.77} & \textbf{3.62} & \textbf{4.41} & \textbf{3.83} & \textbf{2.72} & \textbf{2.66} & \textbf{5.24} & \textbf{3.44} \\
\multicolumn{9}{@{}l}{\rule{0pt}{3.8ex}\quad\itshape Therapist} \\
\quad Joy & \textbf{2.34} & 0.48 & 1.13 & 0.06 & -1.40 & -0.80 & -1.71 & -0.90 \\
\quad Trust & \textbf{6.98} & \textbf{3.37} & \textbf{3.30} & \textbf{3.78} & \textbf{4.46} & \textbf{4.66} & \textbf{4.80} & \textbf{4.15} \\
\quad Fear & -1.57 & 0.17 & 0.01 & -0.15 & -0.64 & -1.25 & -1.48 & \textbf{2.07} \\
\quad Surprise & 1.73 & 1.26 & 1.90 & 0.74 & 1.17 & 1.23 & \textbf{2.37} & \textbf{3.54} \\
\quad Sadness & -0.31 & \textbf{3.39} & \textbf{3.17} & \textbf{3.03} & \textbf{2.55} & \textbf{2.18} & 1.73 & \textbf{3.62} \\
\quad Disgust & \textbf{-5.19} & \textbf{-2.59} & \textbf{-3.52} & \textbf{-2.05} & \textbf{-3.62} & \textbf{-3.92} & \textbf{-4.66} & \textbf{-1.99} \\
\quad Anger & \textbf{-2.84} & -0.88 & -1.55 & -0.98 & -0.81 & -1.54 & \textbf{-3.53} & -0.06 \\
\quad Anticipation & \textbf{5.76} & \textbf{4.21} & \textbf{3.16} & \textbf{3.47} & \textbf{4.39} & \textbf{5.59} & \textbf{4.79} & \textbf{4.56} \\
\midrule
\multicolumn{9}{@{}l}{\textbf{Generalized Anxiety Disorder (GAD)}} \\
\multicolumn{9}{@{}l}{\quad\itshape Patient} \\
\quad Joy & \textbf{3.09} & 0.62 & \textbf{-2.30} & -1.87 & \textbf{-2.05} & \textbf{-2.04} & \textbf{-2.96} & -1.72 \\
\quad Trust & \textbf{5.41} & \textbf{2.37} & \textbf{2.37} & \textbf{2.46} & \textbf{3.05} & 1.07 & \textbf{3.00} & \textbf{2.73} \\
\quad Fear & \textbf{-2.40} & 1.56 & \textbf{3.54} & \textbf{2.09} & \textbf{2.29} & \textbf{2.46} & 0.73 & \textbf{3.09} \\
\quad Surprise & 1.60 & \textbf{2.67} & \textbf{3.18} & \textbf{4.05} & \textbf{3.03} & \textbf{2.77} & \textbf{3.74} & \textbf{4.51} \\
\quad Sadness & 0.30 & \textbf{2.30} & \textbf{2.85} & \textbf{2.97} & 1.89 & \textbf{3.51} & 1.73 & \textbf{2.94} \\
\quad Disgust & \textbf{-3.93} & \textbf{-2.75} & \textbf{-3.16} & \textbf{-3.64} & \textbf{-2.80} & \textbf{-3.23} & \textbf{-3.34} & -1.40 \\
\quad Anger & \textbf{-2.50} & -1.05 & -0.48 & 0.96 & 0.05 & 0.19 & -0.94 & 0.46 \\
\quad Anticipation & \textbf{5.77} & \textbf{4.89} & \textbf{4.38} & \textbf{4.74} & \textbf{4.58} & \textbf{4.09} & \textbf{4.87} & \textbf{4.54} \\
\multicolumn{9}{@{}l}{\rule{0pt}{3.8ex}\quad\itshape Therapist} \\
\quad Joy & \textbf{2.34} & 0.32 & -1.90 & -1.69 & \textbf{-2.15} & \textbf{-2.06} & -1.66 & \textbf{-2.11} \\
\quad Trust & \textbf{6.98} & \textbf{3.97} & \textbf{3.05} & \textbf{4.96} & \textbf{4.12} & \textbf{4.39} & \textbf{5.24} & \textbf{5.42} \\
\quad Fear & -1.57 & 0.07 & \textbf{3.07} & 1.38 & 1.65 & -0.44 & -1.10 & \textbf{2.09} \\
\quad Surprise & 1.73 & 1.55 & \textbf{3.92} & \textbf{2.59} & \textbf{2.71} & \textbf{2.97} & 1.87 & \textbf{2.96} \\
\quad Sadness & -0.31 & 1.45 & \textbf{2.09} & 1.25 & 1.26 & 1.29 & -0.23 & 1.33 \\
\quad Disgust & \textbf{-5.19} & \textbf{-4.51} & \textbf{-4.10} & \textbf{-4.71} & \textbf{-4.86} & \textbf{-4.57} & \textbf{-5.88} & \textbf{-3.76} \\
\quad Anger & \textbf{-2.84} & \textbf{-2.24} & -1.24 & -1.43 & -1.27 & -1.61 & \textbf{-4.44} & -1.01 \\
\quad Anticipation & \textbf{5.76} & \textbf{4.72} & \textbf{4.69} & \textbf{5.09} & \textbf{6.49} & \textbf{6.18} & \textbf{7.33} & \textbf{4.82} \\
\midrule
\multicolumn{9}{@{}l}{\textbf{Borderline Personality Disorder (BPD)}} \\
\multicolumn{9}{@{}l}{\quad\itshape Patient} \\
\quad Joy & \textbf{3.09} & -1.18 & 0.06 & -0.37 & \textbf{-2.31} & \textbf{-2.06} & \textbf{-2.20} & \textbf{-2.19} \\
\quad Trust & \textbf{5.41} & \textbf{2.03} & 1.35 & 0.81 & -0.08 & -0.93 & \textbf{2.11} & \textbf{2.76} \\
\quad Fear & \textbf{-2.40} & \textbf{2.37} & \textbf{3.55} & \textbf{3.35} & \textbf{2.03} & \textbf{2.60} & \textbf{3.30} & \textbf{3.16} \\
\quad Surprise & 1.60 & \textbf{2.41} & \textbf{3.95} & \textbf{3.46} & 1.83 & \textbf{2.54} & \textbf{3.16} & \textbf{4.01} \\
\quad Sadness & 0.30 & \textbf{4.12} & \textbf{4.71} & \textbf{3.57} & \textbf{4.66} & \textbf{4.92} & \textbf{3.81} & \textbf{3.54} \\
\quad Disgust & \textbf{-3.93} & -0.28 & 0.89 & -0.45 & -0.07 & 0.08 & -0.83 & 1.15 \\
\quad Anger & \textbf{-2.50} & \textbf{2.15} & 1.60 & \textbf{2.23} & \textbf{3.17} & \textbf{2.85} & 1.61 & \textbf{2.52} \\
\quad Anticipation & \textbf{5.77} & \textbf{2.13} & \textbf{3.96} & \textbf{2.98} & \textbf{2.37} & \textbf{3.05} & \textbf{3.71} & \textbf{2.66} \\
\multicolumn{9}{@{}l}{\rule{0pt}{3.8ex}\quad\itshape Therapist} \\
\quad Joy & \textbf{2.34} & -1.42 & -0.89 & -1.80 & \textbf{-2.11} & -1.17 & \textbf{-3.01} & \textbf{-2.56} \\
\quad Trust & \textbf{6.98} & \textbf{3.73} & \textbf{2.16} & 1.79 & \textbf{2.56} & \textbf{2.36} & \textbf{4.69} & \textbf{3.31} \\
\quad Fear & -1.57 & 1.56 & \textbf{4.08} & \textbf{4.24} & 1.04 & 1.45 & 1.55 & \textbf{3.26} \\
\quad Surprise & 1.73 & 1.28 & \textbf{2.28} & \textbf{2.77} & \textbf{3.00} & \textbf{2.53} & \textbf{3.93} & \textbf{4.00} \\
\quad Sadness & -0.31 & \textbf{3.66} & \textbf{4.38} & \textbf{4.15} & \textbf{4.04} & \textbf{2.80} & \textbf{3.29} & \textbf{3.75} \\
\quad Disgust & \textbf{-5.19} & \textbf{-2.09} & -1.80 & \textbf{-2.26} & \textbf{-3.41} & \textbf{-3.00} & \textbf{-3.72} & -1.02 \\
\quad Anger & \textbf{-2.84} & 1.28 & 0.14 & 1.13 & 0.96 & 1.15 & -0.89 & \textbf{2.21} \\
\quad Anticipation & \textbf{5.76} & \textbf{3.27} & \textbf{3.44} & \textbf{3.07} & \textbf{4.70} & \textbf{5.34} & \textbf{5.37} & \textbf{3.09} \\
\end{longtable}
\endgroup

\end{document}